\documentclass{article}
\usepackage{spconf,amsmath,graphicx}
\usepackage{booktabs}
\usepackage{algorithm}
\usepackage{lineno,hyperref}
\usepackage{setspace}
\usepackage{algorithm}
\usepackage{cite}
\usepackage{algpseudocode}
\usepackage{amsmath}
\usepackage{graphics}
\usepackage{epsfig}

\title{Learning Data Augmentation Policies Using Augmented Random Search}

\name{Mingyang Geng$^{1}$,
	Kele Xu$^{1}$,
	Bo Ding$^{1}$,
	Huaimin Wang$^{1}$
	Lei Zhang$^{2}$
}

\address{$^{1}$National University of Defense Technology, Changsha, China \\
$^{2}$ Northeastern University, Shenyang, China\\ 
}

\begin{document}
%
\maketitle
\begin{abstract}

Previous attempts for data augmentation are designed manually, and the augmentation policies are dataset-specific.
Recently, an automatic data augmentation approach, named AutoAugment, is proposed using reinforcement learning.
AutoAugment searches for the augmentation polices in the discrete search space, which may lead to a sub-optimal solution.
In this paper, we employ the Augmented Random Search method (ARS) to improve the performance of AutoAugment. 
Our key contribution is to change the discrete search space to continuous space, which will improve the searching performance and maintain the diversities between sub-policies.
With the proposed method, state-of-the-art accuracies are achieved on CIFAR-10, CIFAR-100, and ImageNet (without additional data). Our code is available at
https://github.com/gmy2013/ARS-Aug.

\end{abstract}
\section{Introduction}
\label{sec:intro}
Neural networks are prone to overfit when the labeled data is limited. Regularization is one of the key components to prevent overfitting in the training of deep neural network.
Data augmentation serves as a type of regularization when training neural networks, and it can greatly reduce the change of overfitting. By generating artificial training data via label-preserving transformations of existing training samples,
data augmentation maintains the ability to increase both the amount and diversity of data \cite{krizhevsky2012imagenet, Simard2003Best, baird2012structured}.
Recently, AutoAugment \cite{cubuk2018autoaugment} has been proposed to automatic search for better data augmentation approach that can incorporate invariance and generalize well across different models and datasets. 
AutoAugment enriches the diversity of each policy by introducing a probability and magnitude for each operation. 
And it treats the problem of finding the best augmentation policy as a discrete search task and achieves state-of-the-art accuracy on CIFAR-10, CIFAR-100, SVNH, and ImageNet (without additional data).
However, the probability and magnitude are divided in a discrete space, which may lead to a sub-optimal solution. 
The discrete search task is a common problem in the field of reinforcement learning. 
The search algorithm (an RNN controller and the Proximal Policy Optimization algorithm) chosen by AutoAugment is one of many available search algorithms to find the potential policies. 
However, the final policy found by AutoAugment may have some disadvantages because it contains some sub-policies which are rarely applied in practice as the probability of an operation is too small.
So we suppose that the rarely-used sub-policies could be substituted to get better performance if better search algorithms can be deployed.  

Recently, an Augmented Random Search method (ARS) \cite{mania2018simple}
has shown its immense potential in dealing with continuous control problems. 
In particular, ARS has been proved to match or exceed state-of-the-art sample efficiency on MuJoCo locomotion benchmarks \cite{brockman2016openai,todorov2012mujoco}. In concrete, it is 15 times more computationally efficient than evolution strategy (ES) \cite{salimans2017evolution}, which is the fastest competing method. 
Taking advantage of the high computational efficiency of ARS, we can explore the large policy space more adequately over many random seeds and different choices of hyper-parameters. 
Naturally, with the aim of finding better-augmented policies, we explore to apply ARS to the policy search problem.

In more detail, we aim to substitute the discrete search space with a continuous space while maintaining the efficiency of the search procedure. 
To achieve this goal, we first apply a sigmoid function to normalize the output. 
Then, the normalized output is divided into three categories: operation, probability, and magnitude. 
In the implementation, each policy expresses more accurate states than those generated by Autoaugment because of the continuous policy space. 
With the proposed search approach,  state-of-the-art accuracies have been achieved on the datasets including CIFAR-10, CIFAR-100, and ImageNet (without additional data). 
On CIFAR-10, we achieve an error rate of 1.26$\%$, which is 0.22$\%$ better than the state-of-the-art \cite{cubuk2018autoaugment}. 
On CIFAR-100, we improve the accuracy of AutoAugment from 10.67$\%$ to 10.24$\%$. On ImageNet, we achieve a Top-1 accuracy of 83.88$\%$.

The followings are organized as: Section 2 describes the relationship between our work and previous related work. Section 3 presents the proposed approach, while Section 4 presents the quantitative comparison results. The conclusion is drawn in Section 5.

\section{Releted work}
\label{sec:pagestyle}
Data augmentation is widely-used for visual recognition task, while the policies augmentation are designed manually. Presently, the popular data augmentation approaches include: (1) geometric transformation, such as scale, shifting, rotation, flip, affine transformation (such as elastic distortions \cite{Simard2003Best}). (2) Sample crop and interpolation, such as random Eraser \cite{zhong2017random}, Mixup \cite{zhang2017mixup}. (3) Sample synthetic using Generative Adversarial Neural Networks \cite{goodfellow2014generative}. In \cite{devries2017dataset}, the author proposed to deploy the augmentation in feature space.
All of these manual augmentation techniques are a form of label-preserving data augmentation, which relied on the heavy interaction from the knowledge expert.
AutoAugment is proposed in \cite{cubuk2018autoaugment} to find the policies from the data in an automatic manner. This paper aims to improve the performance of AutoAugment by substituting the discrete search space with the continuous one.

\section{Method}
\label{sec:pagestyle}

We follow the policy definition of AutoAugment: each sub-policy consisting of two image operations to be applied in sequence; each operation is also associated with two hyper-parameters: 1) the probability of applying the operation, and 2) the magnitude of the operation \cite{cubuk2018autoaugment}.  The problem of augmentation policy search can be formulated as a continuous search problem. Notably, the operations are applied in the specified order and the search space for the two hyper-parameters is continuous, which guarantees a diverse sample search processing.

\begin{algorithm}[h] 
\caption{Augmented Random Search for Data Augmentation} 
\label{alg::conjugateGradient} 
\begin{algorithmic}[1] 
\Require 
step-size $\alpha$, number of directions sampled per iteration N, standard deviation
of the exploration noise $\nu$, number of top-performing directions to use $b$.

\State $M_{0}=0\in R^{1\times 30}$,$j=0$; 
\Repeat 
\State Sample $\delta_{1}$,$\delta_{2}$,$\dots$,$\delta_{N}$ in $R^{1\times 30}$ with i.i.d. standard normal entries; 
\State Collect $2N$ outputs by disturbing the $2N$ policies,

\begin{equation}
\left\{
             \begin{array}{lr}
             \pi_{j,k,+}=Sigmoid(M_{j}+ \nu\delta_{k}), &  \\
             
             \pi_{j,k,-}=Sigmoid(M_{j}- \nu\delta_{k}), &  
             \end{array}
\right.
\end{equation}

for $k \in \left\{ 1,2,\dots,N \right\}$ 

\State Transfer each output $\pi_{j,k}$ into five sub-policies based on the type (operation, probability, magnitude) of each dimension, and concatenate them into one single policy.

\State For each generated policy, train a child model using the policy and get the corresponding reward on the test set.

\State Sort the directions $\delta_{k}$ by $max \left\{ r(\pi_{j,k,+}),r(\pi_{j,k,-}) \right\}$, denote by $\delta_{k}$ the k-th largest direction, and by $\pi_{j,k,+}$ and $\pi_{j,k,-}$ the corresponding policies. 

\State Make the update step: 

\begin{align}
M_{j+1}=M_{j}+\frac{\alpha}{b\sigma_ {R}}\sum_{k=1}^{b}\left[r\left(\pi_{j,k,+})-r(\pi_{j,k,-}\right) \right]\delta_{k}
\end{align}
where $\sigma_ {R}$ is the standard deviation of the $2b$ rewards used in the update step.

\State $j=j+1$
\Until{satisfy the ending condition} 
\end{algorithmic} 
\end{algorithm}

We denote our method as ARS-Aug, which build on successful heuristics employed in deep reinforcement learning. The idea of ARS-Aug is to search the best policy directly on the sphere in parameter space. In concrete, it collects the reward signals on a series of directions in parameter space and then optimizes the step along each direction to form the best policy. The reward signal is obtained from the generalization accuracy of a ``child model" (a small neural network which decreases the training time). The data parallel training across multiple GPUs using Ray \cite{moritz2017ray} is exploited to speed up the ``child model" training process and collect more reward signals. For optimizing the steps on each direction to form the final best policy, ARS-Aug updates each perturbation direction $\delta$ by the difference of the rewards $r(\pi_{j,k,+})$ and $r(\pi_{j,k,-})$. This function quantifies the step to move in a certain direction. In addition, we improve the updating process by discarding the computation of the update steps on the directions that yield the least improvement of the reward. This mechanism can guarantee that the update steps are an average over directions that obtained high rewards.  

In order to collect the reward signals, we need to transfer the output of a policy $\pi_{j,k}$ to the data augmentation policy. The output $\pi_{j,k}$ is generated by disturbance on the policy and a sigmoid function for normalizing. The output $\pi_{j,k}$ is a 30-dimensional vector, which needs to be transferred to five sub-polices, each with two operations, and each operation requiring an operation type, magnitude, and probability. For details of the transferring process, we first split the 30-dimensional vector into ten 3-dimensional vectors. Then, for each 3-dimensional vector $a$, the three dimensions stand for operation type ($a[0]$), probability ($a[1]$) and magnitude ($a[2]$) respectively. For the operation type, we discrete the output space by dividing the interval [0,1] into 16 parts, and then map the value to the identifier of the sub-interval $a[0]/(1/16)$. The possibility of each operation is directly represented as the second dimension $a[1]$. Similarly, the magnitude of the operation is transferred as $a[2]\times (Operation_{max}-Operation_{min})+ Operation_{min}$.

\section{Experimental Results}

In this section, we evaluate the performance of ARS-Aug on the CIFAR-10 \cite{krizhevsky2009learning}, CIFAR-100 \cite{krizhevsky2009learning}, and ImageNet \cite{deng2009imagenet} datasets. 
In our experiments, ARS-Aug is implemented with a parallel version using the Python library Ray \cite{moritz2017ray}. A shared noise table storing independent standard normal entries is used in order to avoid the computational bottleneck of communicating perturbations $\delta$. This will guarantee that the workers can communicate indices through the shared noise table. We also set the random seeds for the generators of all the workers. The random seeds are distinct from each other to get a diverse sample efficiency. We repeated the training process 100 times with different random seeds and hyper-parameters for a thorough search over the policy space. The random seeds are sampled uniformly from the interval [0,1000) and are then fixed.

\subsection{CIFAR-10 and CIFAR-100 Results}

As CIFAR-10 and CIFAR-100 have a close data distribution, we aim to find an augmentation policy which can suit for both the two datasets. Considering that the search process needs to take the child model’s accuracy as the reward signal, we establish a reduced CIFAR-10 dataset to decrease the training time, 4000 examples are randomly selected to generate the smaller dataset to decrease the training time. However, the validation process is stochastic due to the process of randomly choosing sub-policies and each operation’s applied probability. To find a suitable number of epochs per sub-policy for ARS-Aug to be effective, we conduct a series of experiments to fix the most approximate value. We find that 120 epochs are suitable for ARS-Aug to train the ``child model" with five sub-policies. In other words, the training time can make the models fully benefit from all of the sub-policies. In addition, we also fix the training epochs for the datasets (e.g., 1800 epochs for Shake-Shake on CIFAR-10, and 270 epochs for ResNet-50 on ImageNet).

We now introduce the details for ARS-Aug to find the best augmentation policy. For the child model architecture, we use small Wide-ResNet-40-2 model, and train for 120 epochs. The use of a small Wide-ResNet is for computational efficiency. We follow the experimental settings \cite{loshchilov2016sgdr}: a weight decay of $10^{-4}$, a learning rate of $0.01$, and a cosine learning decay with one annealing cycle.

It is worthwhile to note that: in order to make full use of the advantages of augmented policies, the augmented policy is applied in addition to the standard baseline pre-processing: on one image, we first apply the baseline augmentation provided by the existing baseline methods, then apply the augmented policy, then apply cutout.

On CIFAR-10, the operations of the policies found by ARS-Aug have no main difference with those of AutoAugment. However, the probability of each operation has been optimized since there does not exist the values which are close to zero. For example, the ''Invert" operation does not appear in the concatenated policy, which is different from that of AutoAugment. This will make room for more meaningful operations and increase the diversity of the whole policy. In addition, the value of magnitude is more accurate (two decimal places) than that of AutoAugment (one decimal), which gives a more precise measurement of the influence brought by each operation. 

The importance of diversity in Augmented policies has been demonstrated in AutoAugment. The hypothesis that more sub-policies will improve the generalization accuracy has been validated \cite{cubuk2018autoaugment}, and the validation accuracy improves with more sub-policies up to about 25 sub-policies. Therefore, we concatenate 25 sub-policies and form a single policy to train on the full datasets.

We now show the advantage of policies found by ARS-Aug on CIFAR-10. We choose six neural network architectures to make a quantitative comparison with AutoAugment. In order to guarantee a fair comparison, we first find the most approximate hyper-parameters for weight decay, and learning rate that give the best validation set accuracy with baseline augmentation. All the other implemented details are the same as reported in the papers which introduce the corresponding models \cite{zagoruyko2016wide, gastaldi2017shake, yamada2018shakedrop}. As shown from Table 1, the test set accuracies of ARS-Aug beat AutoAugment on all the test models. Additionally, we achieve an error rate of 1.26$\%$ with the ShakeDrop \cite{yamada2018shakedrop} model, which is 0.22$\%$ better than AutoAugment.        

\begin{table}[htbp] 
\centering
 \caption{\label{tab:test}Test set error rates ($\%$) on CIFAR-10. Lower is better.} 
 \begin{tabular}{ccc} 
  \toprule 
  Model  & AutoAugment & ARS-Aug  \\ 
  \midrule 
 Wide-ResNet-28-10  &  2.68 & 2.33 \\
  Shake-Shake (26 2$\times$32d)  & 2.47 & 2.14 \\
 Shake-Shake (26 2$\times$96d)  & 1.99 & 1.68 \\
 Shake-Shake (26 2$\times$112d)  & 1.89 & 1.59 \\
 AmoebaNet-B (6,128)  & 1.75 & 1.49\\
 PyramidNet + ShakeDrop  & 1.48 & \textbf{1.26} \\
  \bottomrule 
 \end{tabular} 
\end{table}

We also train three models on CIFAR-100 with the same policy found by ARS-Aug on reduced CIFAR-10; The results are shown in Table 2. Taking advantage of the sampling efficiency of ARS-Aug, it outperforms AutoAugment $0.43\%$ on the error rate. 

\begin{table}[htbp] 
\centering
 \caption{\label{tab:cifar100}Test set error rates ($\%$) on CIFAR-100. Lower is better.} 
 \begin{tabular}{ccc} 
  \toprule 
  Model & AutoAugment & ARS-Aug  \\ 
  \midrule 
 Wide-ResNet-28-10  &  17.09 & 16.64 \\
 Shake-Shake (26 2$\times$96d)  & 14.28 & 13.86 \\
 PyramidNet+ShakeDrop  & 10.67 & \textbf{10.24} \\
  \bottomrule 
 \end{tabular} 
\end{table}

\begin{figure*}[h]
	\centering	
	\includegraphics[width=0.9\textwidth]{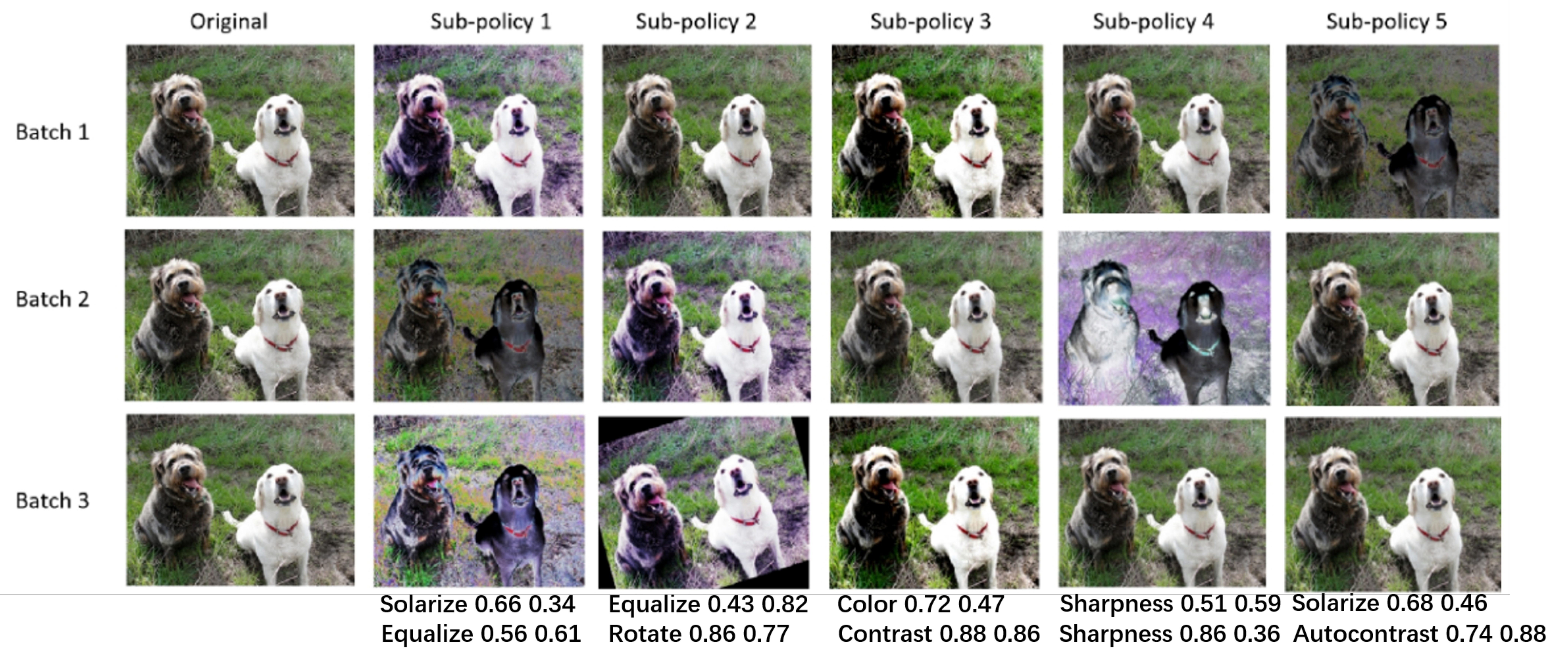}\\
	\caption{The policies learned by ARS-Aug for ImageNet. The policies can be used to generate augmented data given an original image used to train a neural network. We visualize the stochastic in applying the sub-policies by showing how one example image can be transformed differently in different mini-batches. As it can be seen, most of the policies found on ImageNet used color-based transformations.}
	\label{stage}
\end{figure*}

\subsection{ImageNet Results}

Similarly to the above experiments, we apply the same method on ImageNet to find the best-augmented policies. All the implemented details follow those of AutoAugment. The best policies found on ImageNet mainly focus on color-based and Rotation transformation, which have some similarity with those found on CIFAR-10. We then concatenate the best five sub-policies for ImageNet training with the same details of AutoAugment. From Table 3, we can see that the accuracies on all the models are optimized. To our best knowledge, there only exists a better result of $14.6\%$ Top-1 error rate \cite{mahajan2018exploring}, which takes advantage of a large amount of weakly labeled extra data. An illustrative example of selected policies from ARS-Aug is visualized in Fig. 1.

\begin{table}[htbp] 
\centering
 \caption{\label{tab:imagenet}Validation set Top-1 / Top-5 error rates ($\%$) on ImageNet. } 
 \begin{tabular}{ccc} 
  \toprule 
  Model  & AutoAugment & ARS-Aug  \\ 
  \midrule 
 ResNet-50  &  22.37/6.18 & 22.09/5.98  \\
 ResNet-200  & 20.00/4.99 & 19.76/4.67  \\
 AmoebaNet-B (6,190) & 17.25/3.78 & 16.88/3.47 \\
 AmoebaNet-C (6,228) & 16.46/3.52 & \textbf{16.12/3.28} \\
  \bottomrule 
 \end{tabular} 
\end{table}

\section{Conclusion}
\label{sec:copyright}

We have proposed an augmented random search method ARS-Aug for searching better data augmentation policies compared with AutoAugment. The discrete search space of AutoAugment has been changed to a continuous space to improve the searching performance. By fully taking advantage of the higher sample efficiency of ARS, ARS-Aug can find better policies for data augmentation and achieve state-of-the-art accuracy on CIFAR-10, CIFAR-100, and ImageNet (without additional data).  Our work still has some limitations. For example, the datasets we select are limited to the vision domain. Therefore, we consider our future work to apply our automatic augmentation approach to the audio/speech domain and try other search strategies to improve the performance and efficiency.

\section{ACKNOWLEDGMENT}
This work is partially supported by the National Natural Science Foundation of China (nos. 61751208)). The authors would like to acknowledge useful discussions and helpful suggestions from Zhifeng Gao (Microsoft Asia).

\bibliographystyle{IEEEtran}
\bibliography{main.bib}

\end{document}